\documentclass[sigconf]{acmart}

\usepackage{latexsym}
\usepackage{url}
\usepackage{wrapfig}
\usepackage{lipsum,caption}
\usepackage{amsmath}
\usepackage{multirow}
\usepackage{graphicx}
\usepackage{color}
\usepackage[ruled,vlined]{algorithm2e}
\renewcommand\footnotetextcopyrightpermission[1]{}
\usepackage{xspace}
\usepackage{color,soul}
\usepackage[textsize=scriptsize]{todonotes}

\usepackage{physics}
\usepackage{tcolorbox}
\usepackage{xcolor}
\usepackage{amssymb}
\usepackage{comment}
\usepackage{booktabs}
\usepackage{longtable}
\usepackage{enumitem}
\usepackage{todonotes}
\RequirePackage{colortbl}

\definecolor{airforceblue}{rgb}{0.36, 0.54, 0.66}
\definecolor{amaranth}{rgb}{0.9, 0.17, 0.31}
\definecolor{applegreen}{rgb}{0.55, 0.71, 0.0}
\definecolor{alizarin}{rgb}{0.82, 0.1, 0.26}
\definecolor{azure}{rgb}{0.0, 0.5, 1.0}
\definecolor{cadmiumgreen}{rgb}{0.0, 0.42, 0.24}
\newcommand\BibTeX{B\textsc{ib}\TeX}

\AtBeginDocument{%
  \providecommand\BibTeX{{%
    \normalfont B\kern-0.5em{\scshape i\kern-0.25em b}\kern-0.8em\TeX}}}

\copyrightyear{2020} 
\acmYear{2020} 
\setcopyright{acmcopyright}
\acmConference[CoDS COMAD 2020]{7th ACM IKDD CoDS and 25th COMAD}{January 5--7, 2020}{Hyderabad, India}
\acmBooktitle{7th ACM IKDD CoDS and 25th COMAD (CoDS COMAD 2020), January 5--7, 2020, Hyderabad, India}
\acmPrice{15.00}
\acmDOI{10.1145/3371158.3371194}
\acmISBN{978-1-4503-7738-6/20/01}

\begin{document}
\title{Weakly-Supervised Deep Learning for Domain Invariant Sentiment Classification}

\author{Pratik Kayal}
\affiliation{
  \institution{Indian Institute of Technology Gandhinagar}
  \city{Gujarat}
  \state{India}
}

\author{Mayank Singh}
\affiliation{%
  \institution{Indian Institute of Technology Gandhinagar}
 \city{Gujarat}
  \country{India}}
\author{Pawan Goyal}
\affiliation{%
  \institution{Indian Institute of Technology Kharagpur}
  \city{West Bengal}
  \country{India}
}

\begin{abstract}
The task of learning a sentiment classification model that adapts well to any target domain, different from the source domain, is a challenging problem. Majority of the existing approaches focus on learning a common representation by leveraging both source and target data during training. In this paper, we introduce a two-stage training procedure that leverages weakly supervised datasets for developing simple lift-and-shift-based predictive models without being exposed to the target domain during the training phase. Experimental results show that transfer with weak supervision from a source domain to various target domains provides performance very close to that obtained via supervised training on the target domain itself.
\end{abstract}



\maketitle

\section{Introduction}
Sentiment analysis is the practice of applying natural language processing and machine learning techniques to examine the polarity (sentiment) of subjective information from text. With the advancement in the internet infrastructure and cheaper Web services, a large volume of opinionated sentences is produced and consumed by users. The majority of the volume is available in social media, blogs, online retail shops, and discussion forums. Primarily, this user-generated content evaluates the utility and quality of products and their components such as laptops, mobile phones and books, and services such as restaurants, hotels, and events. As of April 2013, 90\% of the customers' purchase decisions are dependent on online reviews ~\citep{peng2014seller}. However, automatic sentiment classification is a challenging problem due to several factors such as the unavailability of the trained dataset~\citep{dasgupta2009mine}, multilingualism~\citep{hogenboom2014multi}, bias~\citep{khan2016sentiment}, etc. We describe relevant past work and our contributions in the following sections.  

\begin{figure}[b]
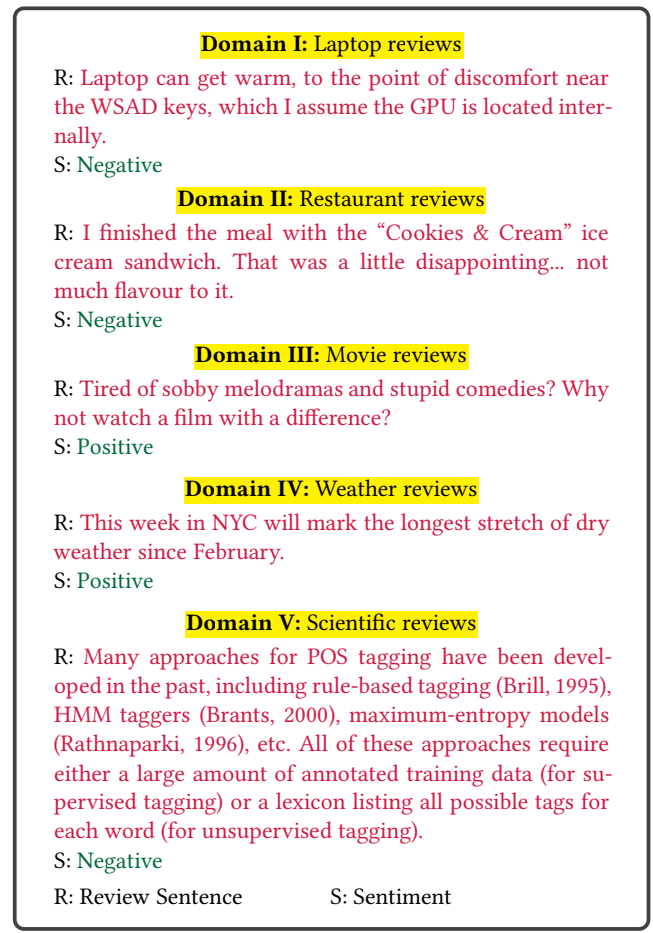

    \centering
    \begin{tcolorbox}[colback=white]
\begin{center} 
\hl{\textbf{Domain I:} Laptop reviews}\\
\end{center}
\textsc{R}: \textcolor{alizarin}{Laptop can get warm, to the point of discomfort near the WSAD keys, which I assume the GPU is located internally.}\\
\textsc{S}: \textcolor{cadmiumgreen}{Negative}
\begin{center} 
\hl{\textbf{Domain II:} Restaurant reviews}\\
\end{center}
\textsc{R}: \textcolor{alizarin}{I finished the meal with the ``Cookies \& Cream'' ice cream sandwich. That was a little disappointing... not much flavour to it.}\\
\textsc{S}: \textcolor{cadmiumgreen}{Negative}
\vspace{0.01cm}
\begin{center} 
\hl{\textbf{Domain III:} Movie reviews}\\
\end{center}
\textsc{R}: \textcolor{alizarin}{Tired of sobby melodramas and stupid comedies? Why not watch a film with a difference?}\\
\textsc{S}: \textcolor{cadmiumgreen}{Positive}
\vspace{0.1cm}
\begin{center} 
\hl{\textbf{Domain IV:} Weather reviews}\\
\end{center}
\textsc{R}: \textcolor{alizarin}{This week in NYC will mark the longest stretch of dry weather since February.}\\
\textsc{S}: \textcolor{cadmiumgreen}{Positive}
\vspace{0.1cm}
\begin{center} 
\hl{\textbf{Domain V:} Scientific reviews}\\
\end{center}
\textsc{R}: \textcolor{alizarin}{Many approaches for POS tagging have been developed in the past, including rule-based tagging (Brill, 1995), HMM taggers (Brants, 2000), maximum-entropy models (Rathnaparki, 1996), etc. All of these approaches require either a large amount of annotated training data (for supervised tagging) or a lexicon listing all possible tags for each word (for unsupervised tagging).}\\
\textsc{S}: \textcolor{cadmiumgreen}{Negative}
\vspace{0.1cm}\\
R: Review Sentence \hspace{1cm} S: Sentiment

\end{tcolorbox}
\caption{Example reviews from five domains.}
    \label{fig:example_problem}
\end{figure}

\subsection{Domain Invariant Sentiment Classification}
Figure \ref{fig:example_problem} shows representative sentences from five unrelated domains. Development of supervised classification models is heavily dependent on labeled datasets, which are rarely available in resource constraint domains. One line of work focuses on creating a general representation for multiple domains based on the co-occurrences of domain-specific and domain-independent features ~\cite{blitzer2007biographies,pan2011domain,yu2016learning,yang2017transfer,bollegala2011using,li2012cross,zhang2014type}. ~\citet{peng2018cross} propose an innovative method to simultaneously extract domain-specific and invariant representations, using both source and target domain labeled data. \citet{qiu2015word} identified domain-specific words to improve cross-domain classification. ~\citet{blitzer2007biographies} propose Structural Correspondence Learning (SCL), which learns a shared feature representation for source and target domains.
~\citet{pan2011domain} propose Spectral Feature Alignment (SFA) to construct the alignment between the source domain and target domain by using the co-occurrence between them, in order to build a bridge between the domains. In general, the above methods leverage both source-target domain pairs for training. Availability of labeled instances in resource-constraint domains is a challenging task. 

As a remedy, recent advancement in transfer learning led to the development of classification models (termed as \textit{`Lift-and-shift models'}) that are trained on a labeled dataset from one domain but can perform significantly better on several other domains~\cite{jiang2007two,aue2005customizing,tan2007novel}. Lift-and-Shift model is a natural extension of single domain sentiment classifier for cross-domain sentiment classification. Here, we pick a trained classifier on a source domain and classify reviews from the target domain without any prior training on the target domain. However, we witness models trained on reviews of a particular domain, do not generally do well when tested on reviews of the unknown and different target domain. The several limitations in the textual domain transfer primarily exist due to out-of-the-vocabulary tokens ~\cite{pan2010cross}, stylistic variations ~\cite{li2018delete}, non-generalizable features ~\cite{jiang2007two}, etc. For example, ~\citet{crammer2007learning} assumed that the distributions of multiple sources are the same, but the labelings of the data from different sources may be different from each other. 

\subsection{Weakly Labelled Review Datasets}
Majority of e-commerce, travel, and restaurant websites (such as Amazon, Flipkart, Airbnb, etc.) allow customers to submit their reviews along with a rating over a five-point scale. Even though the rating might not directly correlate with the sentiment of the review, but they provide weak signals for estimating sentence polarity~\cite{guan2016weakly,zhao2017weakly}. These weakly supervision rating datasets are termed as \textit{``Weakly Labeled Dataset (WLD)''}. We show that WLDs, in addition to the labeled dataset, produce significant improvements in domain transfer for resource-constraint target domains.

\subsection{Our Contribution}
In this work, we propose a two-stage lift-and-shift training procedure that leverages standard labeled sentences along with polarity signals emerging from weakly labeled review datasets. Informally, the proposed model is trained on a single source domain but predicts sentiments for different target domains. We show that even though, BERT (Bidirectional Encoder Representation from Transformers)~\citep{devlin2018bert} and ELMO~\cite{peters2018deep} achieve state-of-the-art performance and outperform the previous benchmarks for single domain sentiment analysis, they consistently fail in the cross-domain sentiment analysis. Our proposed training mechanism adapts to unknown target domains and even performs better than models that explicitly leverage target domain data.

\section{Problem Statement}
In this paper, we address the cross-domain sentiment classification problem. Given a source domain $D_{src}$ and set of target domains  $\{D_1, D_2, D_3, D_4\dots\}$ where $D_1$, $D_2$, $\dots$, $D_N$ represent $N$ distinct target domains ($D_{tar}$) with $D_{src}\neq D_{tar}$, the task is to train a classifier on labelled $D_{src}$ data with high polarity prediction accuracy for the sentences from the $D_{tar}$.

\section{Datasets}
We use two types of review datasets --- (i) the {\em weakly labeled datasets} and (ii) the {\em fully labeled datasets}. Table~\ref{tab:datasets} details the datasets statistics. 

\subsection{Weakly Labelled Datasets (WLD)}
Weakly labeled review datasets contain weak signals about the polarity of review sentences. In the current scenario, the weak signals are represented by user ratings associated with each review. User ratings are considered as noisy labels and would result in significantly weak classifiers. As user ratings lie between 1--5, with one being the worst and five being the best review, we adopt a simple strategy to assign sentiment labels to these sentences.
\begin{equation*}
  \ \   sentiment = \Bigg\{  \begin{tabular}{c}
  positive,  \  \      if 4 or 5 star rated \\
  negative,  \  \      if 1 or 2 star rated
  \end{tabular}
\end{equation*}

Please note that we do not consider three-star rated reviews.  The current study uses three weakly labeled datasets, Amazon product reviews~\citep{zhao2017weakly,mcauley2015image}, YELP restaurant reviews~\citep{yelpDataset} and IMDB movie reviews~\citep{maas2011learning}, which we henceforth refer to as AWLD (Amazon weakly labeled dataset), YWLD (YELP weakly labeled dataset) and IWLD (IMDB weakly labeled dataset), respectively. The WLDs are easier to collect, are not explicitly labeled for sentiments, but are manual ratings given to reviews complementing the review text.  

\subsection{Fully Labeled Datasets (FLD)}
Fully labeled datasets consist of manually labeled review sentences. The reviews are labeled into three classes --- (i) positive, (ii) negative, or (iii) neutral. For the current study, we only consider reviews associated with positive or negative sentiments. The current study leverages six fully labeled datasets,  (i) Weather sentiment data~\citep{weatherDataset}, (ii) IMDB~\citep{kotzias2015group}, (iii) Yelp~\citep{kotzias2015group}, (iv) Amazon (Cell and Accessory)~\citep{kotzias2015group}, (v) Scientific citation context data~\citep{athar:2011:SS}, and (vi) Amazon (Digital Cameras, Cell Phones and Laptops)~\citep{zhao2017weakly}. 

Table~\ref{tab:datasets} presents salient statistics of the two types of review datasets. The experiments have been reported for those source domains that possess corresponding WLDs; thus, scientific citation and weather datasets are not considered as the source domains due to unavailability of WLDs. The compiled datasets are currently available at \url{https://bit.ly/2EnjsSe}.

\begin{table}[!tbh]
\centering
  
  \resizebox{\hsize}{!}{
  \begin{tabular}{llccc} \toprule
  &Dataset&Abbreviation&Reviews& P/N Ratio\\\toprule
  \multirow{3}{*}{\rotatebox[origin=c]{90}{\centering\textsc{\textbf{WLD}}}}
      & {Amazon} &AWLD& 1.1M & 1.43\\
      & {Yelp} &YWLD&1M &3.95\\ 
      & {IMDB} &IWLD&50,000 &1\\ \hline
  \multirow{6}{*}{\rotatebox[origin=c]{90}{\centering\textsc{\textbf{FLD}}}} 
  & Weather & WEAT& 980 & 0.81\\
  & Amazon$^1$ &ACAD&11,800 &0.88\\ 
  & IMDB &IMDB& 1,000 &1\\
  & Yelp  &YELP& 1,000 &1\\
  & Amazon$^2$ &ADLD& 1,000 &1\\
  & Scientific  &SCCD& 700 &2.96\\
  \bottomrule \hline
\end{tabular}}
\caption{Salient statistics of the datasets. Appropriate abbreviations are added for better readability in further sections. Amazon$^1$ represents the {\em Cell and Accessory} category and Amazon$^2$ represents {\em Digital Cameras, Cell Phones, and Laptops} category. The rightmost column displays the ratio of counts of positive reviews (P) to negative reviews (N).}
\label{tab:datasets}
\end{table}                            

\section{Methodology}
\subsection{The Training Procedure}
As discussed in previous sections, the intuition is to leverage weak signals generated by WLDs to complement the polarity classification training of FLDs. We, therefore, present a two-stage training procedure. In the first stage, the predictive model is pre-trained using WLD data with lower learning rate for few iterations\footnote{In the current paper, a single iteration over all training instances (one epoch).}. The second stage follows a standard training procedure. In the second stage, the predictive model is trained using FLD instances with usually higher learning rate till convergence. Algorithm~\ref{alg:weak} presents a detailed methodology. The training procedure is followed by the testing procedure. Since, the current work focuses on domain invariant sentiment classification, the test dataset domain, in the majority of the cases, is different from the train dataset domain.   

\begin{algorithm}
\KwIn{\\$D_{scr}$: Source domain\;$D_1, D_2, D_3, D_4,\dots$, $D_N$ : $N$ target domains\;}
 \textbf{Pre-train} model on {\em WLD} of {\em source domain} $D_{scr}$ with a low learning rate and in `$n$' epochs, where $n$ is a small integer\;
  \textbf{Train} model on {\em FLD} of {\em source domain} $D_{scr}$\;
  \For{$i = 1;\ i \leq N;\ i = i + 1$}{
    Test the  trained model on $D_i$\;
  }
 \caption{Weakly Supervised training for Domain Generalization}
 \label{alg:weak}
 \end{algorithm}

\subsection{The Model Architecture}
For the current polarity prediction task, we train a standard fully-connected feed-forward network with softmax as activation function and two output perceptrons. We plug this fully connected layer over recently published state-of-the-art natural language embedding models. We experiment with two models, (i) BERT$_{base}$~\citep{devlin2018bert} and (ii) ELMO~\citep{peters2018deep}. These models are trained on the general Wikipedia articles. Our proposed algorithm considers their pre-trained weights as the initialization for basic language understanding and adapts them to the specific polarity knowledge.

\begin{table*}[!tbh]
\centering
\resizebox{\hsize}{!}{%
\begin{tabular}{lccclcclcclcclcclcc}
\hline \toprule
&& \multicolumn{17}{c}{Target domains} \\\cline{3-19}

&& \multicolumn{2}{c}{ACAD}&&\multicolumn{2}{c}{WEAT}  & &\multicolumn{2}{c}{ADLD} &&\multicolumn{2}{c}{IMDB}    && \multicolumn{2}{c}{YELP}    && \multicolumn{2}{c}{SCCD} \\ \cline{3-4}\cline{6-7}\cline{9-10}\cline{12-13}\cline{15-16}\cline{18-19}
  
 &&A(100\%) & \multicolumn{1}{c}{F1} && A(100\%) & \multicolumn{1}{c}{F1} && A(100\%) & \multicolumn{1}{c}{F1} && A(100\%) & \multicolumn{1}{c}{F1} && A(100\%) & \multicolumn{1}{c}{F1} && A(100\%)   &F1           \\
 \multirow{8}{*}{\rotatebox[origin=c]{90}{\textbf{Source domains}}}& ACAD& \textcolor{blue}{82.50}&\textcolor{blue}{0.809}  &&74.60&0.691  &&{87.30}&{0.871}  &&74.00&0.735  &&75.30&0.764  &&71.20&0.809 \\
 & WEAT & \cellcolor[gray]{0.9}69.10&\cellcolor[gray]{0.9}0.570  &\cellcolor[gray]{0.9}&\cellcolor[gray]{0.9}\textcolor{blue}{82.10}&\cellcolor[gray]{0.9}\textcolor{blue}{0.813} &\cellcolor[gray]{0.9} &\cellcolor[gray]{0.9}69.30&\cellcolor[gray]{0.9}0.596  &\cellcolor[gray]{0.9}&\cellcolor[gray]{0.9}72.70&\cellcolor[gray]{0.9}0.682  &\cellcolor[gray]{0.9}&\cellcolor[gray]{0.9}70.70&\cellcolor[gray]{0.9}0.735  &\cellcolor[gray]{0.9}&\cellcolor[gray]{0.9}29.40&\cellcolor[gray]{0.9}0.182        \\
& ADLD& 75.30&0.775  &&68.70&0.687  &&\textcolor{red}{88.00}&\textcolor{red}{0.878}  &&73.30&0.750  &&78.00&0.802  &&69.30&0.791        \\
 &IMDB &\cellcolor[gray]{0.9} 77.30&\cellcolor[gray]{0.9}0.756  &\cellcolor[gray]{0.9}&\cellcolor[gray]{0.9}{74.60}&\cellcolor[gray]{0.9}\textcolor{red}{0.761}  &\cellcolor[gray]{0.9}&\cellcolor[gray]{0.9}86.00&\cellcolor[gray]{0.9}0.847  &\cellcolor[gray]{0.9}&\cellcolor[gray]{0.9}\textcolor{red}{78.70}&\cellcolor[gray]{0.9}\textcolor{red}{0.802}  &\cellcolor[gray]{0.9}&\cellcolor[gray]{0.9}{79.30}&\cellcolor[gray]{0.9}{0.812}  &\cellcolor[gray]{0.9}&\cellcolor[gray]{0.9}70.60&\cellcolor[gray]{0.9}0.802        \\
 &YELP & 76.10&0.744  &&74.60&0.746  &&82.70&0.783  &&74.70&0.740  &&\textcolor{blue}{81.30}&\textcolor{red}{0.823}  &&37.90&0.371        \\
 &SCCD&\cellcolor[gray]{0.9} 47.80&\cellcolor[gray]{0.9}0.647 &\cellcolor[gray]{0.9} &\cellcolor[gray]{0.9}44.80&\cellcolor[gray]{0.9}0.619  &\cellcolor[gray]{0.9}&\cellcolor[gray]{0.9}48.00&\cellcolor[gray]{0.9}0.649  &\cellcolor[gray]{0.9}&\cellcolor[gray]{0.9}52.70&\cellcolor[gray]{0.9}0.687  &\cellcolor[gray]{0.9}&\cellcolor[gray]{0.9}49.30&\cellcolor[gray]{0.9}0.661  &\cellcolor[gray]{0.9}&\cellcolor[gray]{0.9}\textcolor{red}{75.80}&\cellcolor[gray]{0.9}\textcolor{blue}{0.862}        \\\cline{2-19}
 &AWLD & 54.40&0.676  &&46.30&0.625  &&52.70&0.670  &&52.00&0.684  &&53.30&0.679  &&\textcolor{red}{75.80}&\textcolor{blue}{0.862}  \\
 & IWLD    &\cellcolor[gray]{0.9} 67.30&  \cellcolor[gray]{0.9}  0.699&\cellcolor[gray]{0.9}&\cellcolor[gray]{0.9}    59.70&   \cellcolor[gray]{0.9} 0.64&\cellcolor[gray]{0.9}&\cellcolor[gray]{0.9}    79.30&\cellcolor[gray]{0.9}    0.805&\cellcolor[gray]{0.9}&\cellcolor[gray]{0.9}    74.70&\cellcolor[gray]{0.9}    0.729&\cellcolor[gray]{0.9}&\cellcolor[gray]{0.9}    74.00&\cellcolor[gray]{0.9}    0.755&\cellcolor[gray]{0.9}&\cellcolor[gray]{0.9}    66.40&\cellcolor[gray]{0.9}    0.762 \\
& YWLD    & 73.90&    0.765&&    68.70&    0.72&&    84.00&    0.848&&    74.00&    0.78&&    80.00&    \textcolor{blue}{0.824}&&    \textcolor{blue}{77.80}&    \textcolor{red}{0.861}   \\\cline{2-19}
 &  ACAD-AWLD& \cellcolor[gray]{0.9}\textcolor{red}{80.00}&\cellcolor[gray]{0.9}\textcolor{red}{0.794}  &\cellcolor[gray]{0.9}&\cellcolor[gray]{0.9}71.60&\cellcolor[gray]{0.9}0.708  &\cellcolor[gray]{0.9}&\cellcolor[gray]{0.9}80.70&\cellcolor[gray]{0.9}0.803  &\cellcolor[gray]{0.9}&\cellcolor[gray]{0.9}{75.30}&\cellcolor[gray]{0.9}{0.764}  &\cellcolor[gray]{0.9}&\cellcolor[gray]{0.9}73.30&\cellcolor[gray]{0.9}0.762  &\cellcolor[gray]{0.9}&\cellcolor[gray]{0.9}69.90&\cellcolor[gray]{0.9}0.807\\
& IMDB-IWLD   & 75.60&    0.74&&    68.70&    0.704&&    80.70&    0.788&&    \textcolor{blue}{83.30} &    \textcolor{blue}{0.843} &&    74.70&    0.756&&    61.00&    0.691    \\
 &YELP-YWLD&\cellcolor[gray]{0.9}  76.40&\cellcolor[gray]{0.9}    0.755&\cellcolor[gray]{0.9}&\cellcolor[gray]{0.9}    \textcolor{red}{76.10}&\cellcolor[gray]{0.9}    0.742&\cellcolor[gray]{0.9}&\cellcolor[gray]{0.9}    \textcolor{blue}{90.00}&\cellcolor[gray]{0.9} \textcolor{blue}{0.891}&\cellcolor[gray]{0.9}&\cellcolor[gray]{0.9}    73.30&\cellcolor[gray]{0.9}    0.71&\cellcolor[gray]{0.9}&\cellcolor[gray]{0.9}    \textcolor{red}{80.70}&\cellcolor[gray]{0.9}    {0.818}&\cellcolor[gray]{0.9}&\cellcolor[gray]{0.9}    58.30&\cellcolor[gray]{0.9}    0.56   \\
 \bottomrule
\end{tabular}}
\caption{[Color online] Accuracy and F1 scores with ELMO as embeddings. Here, each target domain represents a 15\% held-out data. Blue and red color represent the best and second best values for a given target domain. AWLD, IWLD and YWLD represent training on weakly labeled datasets only. ACAD-AWLD, IMDB-IWLD and YELP-YWLD represents our two-stage training procedure.}
\label{tab:ELMO}
\end{table*}


\begin{table*}[!tbh]
\centering
\resizebox{\textwidth}{!}{%
\begin{tabular}{lccclcclcclcclcclcc}
\hline \toprule
&& \multicolumn{17}{c}{Target Domains} \\\cline{3-19}
&& \multicolumn{2}{c}{ACAD}&&\multicolumn{2}{c}{WEAT}  & &\multicolumn{2}{c}{ADLD} &&\multicolumn{2}{c}{IMDB}    && \multicolumn{2}{c}{YELP}    && \multicolumn{2}{c}{SCCD} \\ \cline{3-4}\cline{6-7}\cline{9-10}\cline{12-13}\cline{15-16}\cline{18-19}
 &&A(100\%) & \multicolumn{1}{c}{F1} && A(100\%) & \multicolumn{1}{c}{F1} && A(100\%) & \multicolumn{1}{c}{F1} && A(100\%) & \multicolumn{1}{c}{F1} && A(100\%) & \multicolumn{1}{c}{F1} && A(100\%)   &F1           \\

  \multirow{12}{*}{\rotatebox[origin=c]{90}{\textbf{Source Domains}}}& 
  ACAD& \textcolor{red}{90.20}& \textcolor{red}{0.898}    && 80.50  & 0.805     & & \textcolor{blue}{95.30}& \textcolor{blue}{0.953}    && 89.30& 0.901     && 88.60& 0.887     && 80.20& 0.876 \\  
 & WEAT &\cellcolor[gray]{0.9} 79.90&\cellcolor[gray]{0.9} 0.766    &\cellcolor[gray]{0.9}&\cellcolor[gray]{0.9} 85.00  &\cellcolor[gray]{0.9} \textcolor{red}{0.848}      &\cellcolor[gray]{0.9}&\cellcolor[gray]{0.9} 86.60&\cellcolor[gray]{0.9} 0.857    &\cellcolor[gray]{0.9}&\cellcolor[gray]{0.9} 87.30&\cellcolor[gray]{0.9} 0.876     &\cellcolor[gray]{0.9}&\cellcolor[gray]{0.9} 85.30&\cellcolor[gray]{0.9} 0.853     &\cellcolor[gray]{0.9}&\cellcolor[gray]{0.9} 67.00&\cellcolor[gray]{0.9} 0.731    \\
& ADLD      & 84.10& 0.819    && 85.00  & 0.827      && \textcolor{red}{95.30}& \textcolor{red}{0.950}    && 84.60& 0.841     && 91.30& 0.909     && 50.80& 0.867    \\
&  IMDB   &\cellcolor[gray]{0.9} 82.30&\cellcolor[gray]{0.9} 0.805    &\cellcolor[gray]{0.9}&\cellcolor[gray]{0.9} \textcolor{blue}{86.50}  & \cellcolor[gray]{0.9}\textcolor{blue}{0.852}      &\cellcolor[gray]{0.9}&\cellcolor[gray]{0.9} 92.60& \cellcolor[gray]{0.9}0.925    &\cellcolor[gray]{0.9}&\cellcolor[gray]{0.9} 91.30&\cellcolor[gray]{0.9} 0.915     &\cellcolor[gray]{0.9}&\cellcolor[gray]{0.9} 87.30& \cellcolor[gray]{0.9}0.875     &\cellcolor[gray]{0.9}&\cellcolor[gray]{0.9} 65.80&\cellcolor[gray]{0.9} 0.867    \\
& YELP   & 81.90& 0.791    && 79.10  & 0.787      && 89.30& 0.887    && 87.30& 0.872     && \textcolor{blue}{95.30}& \textcolor{blue}{0.953}     && 55.00& 0.867  \\
& SCCD     & \cellcolor[gray]{0.9}73.60&\cellcolor[gray]{0.9} 0.747    &\cellcolor[gray]{0.9}&\cellcolor[gray]{0.9} 56.70  & \cellcolor[gray]{0.9}0.658      &\cellcolor[gray]{0.9}& \cellcolor[gray]{0.9}76.60&\cellcolor[gray]{0.9} 0.782    &\cellcolor[gray]{0.9}&\cellcolor[gray]{0.9} 76.00&\cellcolor[gray]{0.9} 0.785     &\cellcolor[gray]{0.9}&\cellcolor[gray]{0.9} 87.30& \cellcolor[gray]{0.9}0.883     &\cellcolor[gray]{0.9}&\cellcolor[gray]{0.9} \textcolor{red}{82.00}& \cellcolor[gray]{0.9}\textcolor{red}{0.883} \\\cline{2-19}
& AWLD    & 78.65& 0.812    && 50.70  & 0.645      && 89.30& 0.898    && 78.60& 0.829    & & 79.30& 0.82      && 77.20& 0.87     \\
& IWLD    & \cellcolor[gray]{0.9}76.00& \cellcolor[gray]{0.9}0.785    &\cellcolor[gray]{0.9}&\cellcolor[gray]{0.9} 73.10  & \cellcolor[gray]{0.9}0.769     & \cellcolor[gray]{0.9}& \cellcolor[gray]{0.9}74.00& \cellcolor[gray]{0.9}0.782   & \cellcolor[gray]{0.9}& \cellcolor[gray]{0.9}84.70& \cellcolor[gray]{0.9}0.862     &\cellcolor[gray]{0.9}& \cellcolor[gray]{0.9}79.30& \cellcolor[gray]{0.9}0.825     &\cellcolor[gray]{0.9}& \cellcolor[gray]{0.9}\textcolor{blue}{82.00}& \cellcolor[gray]{0.9}\textcolor{blue}{0.893}    \\
& YWLD    & 85.40& 0.855    && 76.10  & 0.778      && 94.00& 0.941    && 90.70& 0.912     && 91.30& 0.916     && 76.60& 0.868    \\\cline{2-19}
& ACAD-AWLD     & \cellcolor[gray]{0.9}\textcolor{blue}{90.50}&\cellcolor[gray]{0.9} \textcolor{blue}{0.903}    &\cellcolor[gray]{0.9}& \cellcolor[gray]{0.9}85.00  &\cellcolor[gray]{0.9} 0.843      &\cellcolor[gray]{0.9}& \cellcolor[gray]{0.9}94.00& \cellcolor[gray]{0.9}0.938    &\cellcolor[gray]{0.9}& \cellcolor[gray]{0.9}\textcolor{red}{92.60}& \cellcolor[gray]{0.9}\textcolor{blue}{0.931}     &\cellcolor[gray]{0.9}& \cellcolor[gray]{0.9}92.00& \cellcolor[gray]{0.9}0.921    &\cellcolor[gray]{0.9} &\cellcolor[gray]{0.9} 80.80& \cellcolor[gray]{0.9}0.879    \\
& IMDB-IWLD   & 86.50& 0.857    && \textcolor{red}{85.10}  & 0.844      && 93.30& 0.933   & & \textcolor{blue}{92.70}&\textcolor{red}{0.929}&& 92.70& 0.928     && 76.60& 0.868    \\
& YELP-YWLD   & \cellcolor[gray]{0.9}86.00& \cellcolor[gray]{0.9}0.853    &\cellcolor[gray]{0.9}&\cellcolor[gray]{0.9} 80.60  & \cellcolor[gray]{0.9}0.806      &\cellcolor[gray]{0.9}& \cellcolor[gray]{0.9}\textcolor{blue}{95.30}& \cellcolor[gray]{0.9}\textcolor{blue}{0.953}    &\cellcolor[gray]{0.9}& \cellcolor[gray]{0.9}92.00& \cellcolor[gray]{0.9}0.923     &\cellcolor[gray]{0.9}& \cellcolor[gray]{0.9}\textcolor{red}{95.00}& \cellcolor[gray]{0.9}\textcolor{red}{0.951}     &\cellcolor[gray]{0.9}& \cellcolor[gray]{0.9}71.90& \cellcolor[gray]{0.9}0.868    \\ \bottomrule
\end{tabular}}
\caption{[Color online] Accuracy and F1 scores with BERT as embeddings. Here, each target domain represents a 15\% held-out data. Blue and red color represent the best and second best values for a given target domain. AWLD, IWLD and YWLD represent training on weakly labeled datasets only. ACAD-AWLD, IMDB-IWLD and YELP-YWLD represents our two-stage training procedure.}
\label{tab:BERT Results}
\vspace{-0.2cm}
\end{table*}

\section{Experiments}

\subsection{Preprocessing and Data Preparation}
Each review is tokenized and subjected to standard token and character level filtering such as lower-casing and special character filtering. Also, all neutral reviews are filtered. Next, each FLD dataset is randomly split into two sets --- training, and test. We allocate 85\% of pairs for training. The rest 15\% of review sentences are allocated for the test. Note that, WLDs are not subjected to random splitting.

\subsection{Lift-and-shift Baselines}
We compare our model with several lift-and-shift baselines with no WLD-based pretraining. Lift-and-shift models are a natural extension of single-domain sentiment classification models. Here, we pick a trained classier model on a source domain and predict sentences from the target domain without any prior training on the target domain. The training procedure does not involve the WLD dataset either. 

\subsection{Evaluation Metrics}
We compare our proposed model against standard lift-and-shift baselines (described in the previous section). We leverage standard metrics in sentiment classification --- (i) \textit{Average accuracy score} and (ii) \textit{F1 score} --- for comparing the predicted polarity with the ground-truth polarity.

\subsection{Experimental Settings}
The BERT and ELMO embedding vector size are fixed at 768 and 512 dimensions, respectively. We used a batch size of 64, and binary cross-entropy loss function.  In the case of BERT, we find that the learning rate of 3.00e-5 and 3.00e-8 for training and pretraining phase, respectively, were performing the best. Similarly, in the case of ELMO, the learning rate of 3.00e-03 and 3.00e-06 for training and pretraining phase performed best.

\section{Results and Discussion}
\label{result}
We, first of all, discuss the performance score of EMLO-based model. Table~\ref{tab:ELMO} presents accuracy and F1 scores for ELMO-based classification model. Here, each target domain represents 15\% test data. ELMO baselines performed best when the system is trained and tested on the same domain (showed by diagonal cells in Table~\ref{tab:ELMO}). Only training on WLD datasets produced worst results (see the row with source as AWLD). Our proposed training procedure (for example, ACAD-AWLD) produced marginal performance. We witness similar poor and marginal performances for other WLDs and for our other proposed two-stage experiment settings, respectively. 
We claim that the poorer EMLO's performance gains in cross-domain classification are primarily due to its current limitations in generalizability as compare to other state-of-the-art embeddings generated from transformer-based language models. Next, we experiment with other competitive models. 

Table~\ref{tab:BERT Results} presents accuracy and F1 scores for BERT-based classification model. BERT performed significantly better than ELMO. Our proposed training procedure performed exceptionally well, in some cases even at par with the models that are trained and tested on the same domain. This training procedure not only improves against standard lift-and-shift models but also leads to higher transfer results. Again, training only on WLD datasets produced the worst results. However, the values are better than the corresponding ELMO-based models. Even though, YELP-YWLD performs marginally poor (~0.3\% lower than YELP), it performs significantly better in other domains such as ADLD (6\% higher than YELP) and IMDB (4.7\% higher than YELP). Similar transfer improvements are reported for ACAD-AWLD and IMDB-IWLD. This transfer performance improvement reconfirms the usefulness of two-stage training procedure. As expected, domain transfer from SCCD and WEAT is inferior due to high dissimilarity between SCCD and WEAT and other domains. 

Note that the performance of the models trained on only WLDs is poor compared to the performance of FLDs. This observation suggests that WLDs are itself very noisy and not good enough for the sentiment classification task\cite{pang2002thumbs, pang2005seeing}. Also, the poor performance of same domain weakly labelled dataset on fully labelled data suggests that the star rating is not highly correlated to sentiment.

\section{Conclusion and Future Work}
In this paper, we propose a two-stage training framework for cross-domain sentiment classification. We showcase the utility of combining weak labels and full labels for domain-invariant sentiment analysis. Even though the proposed approach uses more data than the baseline, curating WLD data is an extremely easier task than the curation of FLDs. WLD datasets extend the transfer capabilities to a significantly large extent.  Our experimental results on BERT based model demonstrate the effectiveness of our proposed framework for a wide range of domains.

The primary focus of this paper has been to train more generalizable sentiment classification models using only single domain data without using any target domain signals. The idea of pretraining on weak signals can further be explored by combining various weakly labeled domains.

\bibliographystyle{ACM-Reference-Format}
\bibliography{sample-base}

\end{document}